\documentclass[11pt]{article}
\usepackage{graphicx}
\usepackage[final]{acl}

\usepackage{times}
\usepackage{latexsym}
\usepackage{hyperref}
\usepackage{url}
\usepackage{graphicx}
\usepackage{booktabs}
\usepackage{multirow}
\usepackage[table]{xcolor}
\usepackage{amsmath}
\usepackage[ruled,linesnumbered]{algorithm2e}
\usepackage{makecell}
\usepackage{subcaption}
\usepackage[T1]{fontenc}

\usepackage[utf8]{inputenc}

\usepackage{microtype}

\usepackage{inconsolata}

\usepackage{graphicx}

\title{DVD: A Robust Method for Detecting Variant Contamination in Large Language Model Evaluation}

\author{
  Renzhao Liang$^{1}$\thanks{First author. \href{mailto:liangrzh@buaa.edu.cn}{liangrenzhao@buaa.edu.cn}} \quad
  Jingru Chen$^{2}$ \quad
  Bo Jia$^{3}$ \quad
  Bo Deng$^{1}$ \quad 
  Chenggang Xie$^{1}$ \quad \\
  Yidong Wang$^{2}$ \quad
  Ke Jin$^{1}$ \quad
  Xin Wang$^{2}$ \quad 
  Linfeng Zhang$^{4}$ \quad
  Cunxiang Wang$^{5}$\thanks{Corresponding author. \href{mailto:wangcunxiang303@gmail.com}{wangcunxiang303@gmail.com}} \\ \\
  $^{1}$Beihang University \hspace{0.5cm}
  $^{2}$Peking University \hspace{0.5cm} 
  $^{3}$Beijing University of Posts and Telecommunications \hspace{0.5cm} \\
  $^{4}$Shanghai Jiao Tong University \hspace{0.5cm}
  $^{5}$Tsinghua University
}

\begin{document}

\maketitle
\begin{abstract}
Evaluating large language models (LLMs) is increasingly confounded by \emph{variant contamination}: the training corpus contains semantically equivalent yet lexically or syntactically altered versions of test items. Unlike verbatim leakage, these paraphrased or structurally transformed variants evade existing detectors based on sampling consistency or perplexity, thereby inflating benchmark scores via memorization rather than genuine reasoning. We formalize this problem and introduce \textbf{DVD} (\textbf{D}etection via \textbf{V}ariance of generation \textbf{D}istribution), a single-sample detector that models the local output distribution induced by temperature sampling. Our key insight is that contaminated items trigger alternation between a \emph{memory-adherence} state and a \emph{perturbation-drift} state, yielding abnormally high variance in the synthetic difficulty of low-probability tokens; uncontaminated items remain in drift with comparatively smooth variance. We construct the first benchmark for variant contamination across two domains Omni-MATH and SuperGPQA by generating and filtering semantically equivalent variants, and simulate contamination via fine-tuning models of different scales and architectures (Qwen2.5 and Llama3.1). Across datasets and models, \textbf{DVD} consistently outperforms perplexity-based, Min-$k$\%++, edit-distance (CDD), and embedding-similarity baselines, while exhibiting strong robustness to hyperparameters. Our results establish variance of the generation distribution as a principled and practical fingerprint for detecting variant contamination in LLM evaluation.
\end{abstract}

\section{Introduction}

In recent years, large language models (LLMs) have exhibited explosive growth in capability, demonstrating transformative potential across a wide range of domains \citep{gpt3,gemini,llama,palm,gpt4}. However, their impressive performance relies heavily on massive web-scale corpora, which has brought a long-standing challenge into sharper focus: \emph{data contamination} \citep{balloccu2024leak,li2023open,chang2024survey,cheng2025survey,deng2023investigating,xu2024benchmarking}. Data contamination refers to unintended overlap between training data and evaluation benchmarks, which severely undermines the validity of empirical evaluation \citep{cheng2025survey}. Such overlap can create an illusion of strong generalization and mislead research progress. When contaminated models are deployed in scientific investigations or real-world applications, their latent biases and hidden flaws may lead to incorrect scientific conclusions or catastrophic decisions, ultimately hindering technological advancement \citep{sainz2023nlp}.

\begin{figure}[t]
    \centering
    \includegraphics[width=\linewidth]{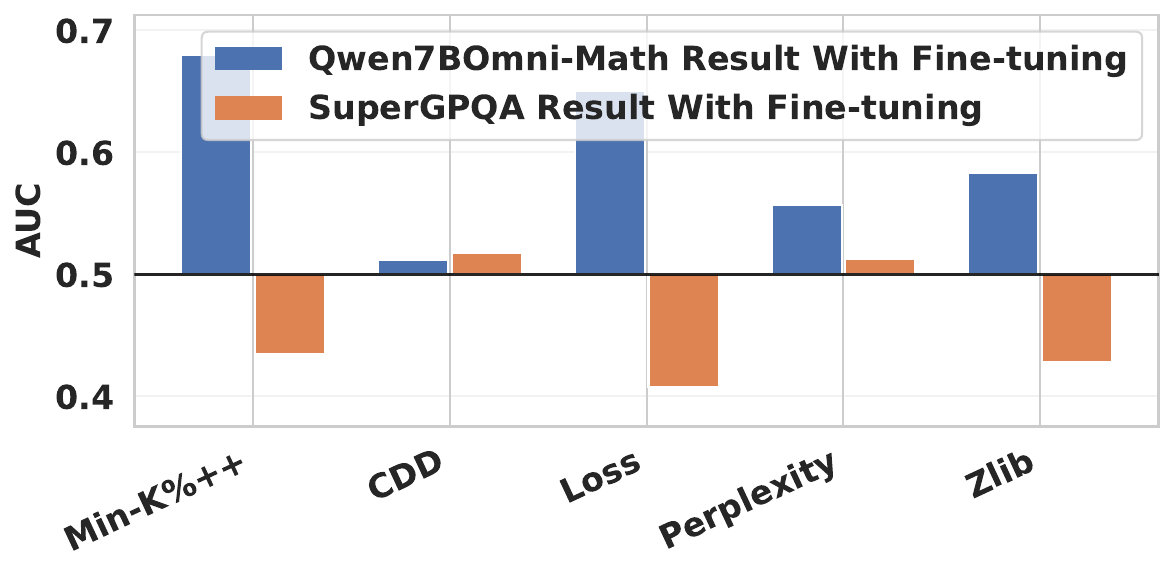}
    \caption{Performance of traditional contamination detection methods on the variant contamination identification task. In the figure, an AUC value below 0.5 indicates that the model’s predictions are inversely correlated with the true labels, while an AUC close to 0.5 suggests performance equivalent to random guessing.}
    \label{fig:intro}
\end{figure}
With the widespread adoption of large-scale data augmentation and synthetic data generation (e.g., GPT-4o), a more subtle and potentially more dangerous form of contamination has gradually emerged, namely \emph{variant contamination}. Variant contamination occurs when training data contains instances that are semantically equivalent to benchmark questions but have been rewritten at the lexical or structural level. Unlike exact duplicates, such variants can evade existing detection methods while still enabling models to effectively "memorize" the answers. To systematically study this phenomenon, we construct a new evaluation benchmark based on Omni-MATH (mathematical reasoning) \citep{gao2024omni} and SuperGPQA (general reasoning) \citep{du2025supergpqa} by generating semantically equivalent variants through controlled transformations. Fine-tuning models of different scales and architectures on these contaminated datasets yields striking results: even when the training data contains only variants (with no exact duplicates), models can still achieve significantly inflated accuracy on the evaluation benchmarks. More importantly, widely used contamination detection methods fail in this setting and exhibit unstable performance (Figure~\ref{fig:intro}). Existing approaches, such as perplexity-based metrics, Min-K\%, and CDD, primarily rely on shallow surface-level features, including token probability distributions, embedding similarity, or surface perplexity patterns. However, prior work has shown that LLMs can be highly sensitive to minor phrasing changes, often leading to substantially different response behaviors \cite{lunardi2025robustness,sclar2023quantifying,zhao2024improving}. Consequently, in the variant contamination scenario, although the questions are semantically equivalent to the benchmarks, their carefully restructured surface forms weaken these shallow cues, making it difficult for existing methods to capture the true behavioral differences exhibited by models. This observation is consistent with our empirical findings.

\begin{figure*}[ht]
    \centering
    \includegraphics[width=\textwidth]{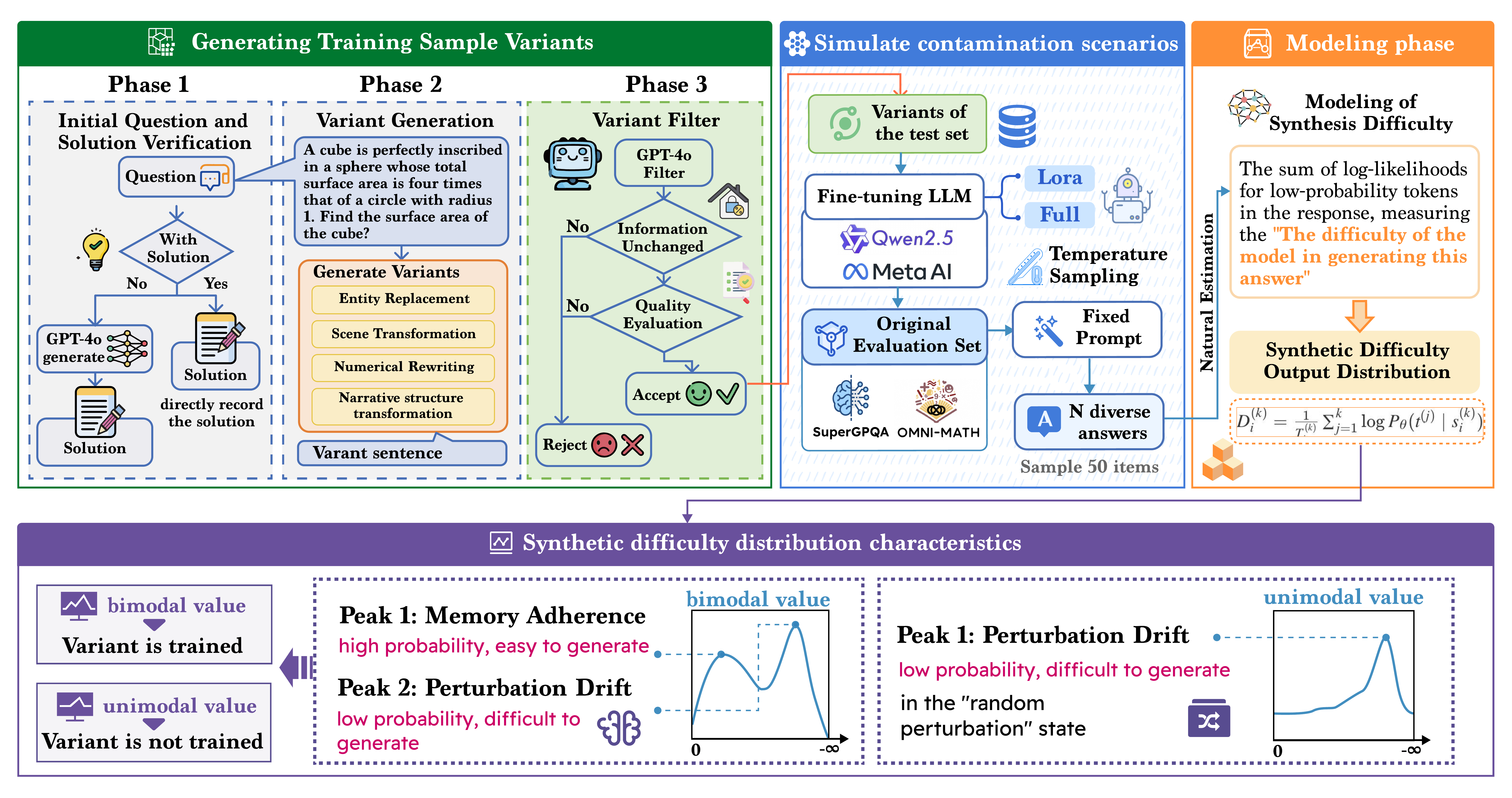}
    \caption{Our Method Pipeline}
    \label{fig:pipeline}
\end{figure*}

To address these limitations, we propose \textbf{DVD} (\textit{Detection via Variance of generation Distribution}), a variant contamination detection method based on the variance of generation distributions. DVD directly characterizes the core behavioral signatures induced by variant contamination by modeling fluctuations in the model's generation distribution across multiple stochastic decoding runs. Unlike existing methods that rely on static surface-level features, DVD focuses on dynamic response patterns in the model's uncertainty space. For uncontaminated questions, genuine reasoning processes typically produce relatively stable and smooth variance in the output distribution. In contrast, for contaminated questions, models frequently alternate between a \emph{high-confidence memorization regime} and a \emph{low-confidence exploratory reasoning regime}, resulting in pronounced distributional differences in the generation variance (Figure~\ref{fig:pipeline}). By exploiting these dynamic behavioral differences, DVD is able to penetrate surface-level reformulations and directly identify contamination-induced anomalies, enabling robust and effective detection in the highly challenging variant contamination setting.

Extensive experiments demonstrate that DVD consistently outperforms baseline methods across datasets, domains, and model scales. For example, on SuperGPQA, DVD improves AUC by up to 0.22 over the strongest baseline (embedding similarity), while maintaining stable performance across model sizes from 1.5B to 32B parameters, as well as across both Qwen and Llama architectures. These results collectively establish DVD as a robust and efficient solution to the overlooked yet critical problem of variant contamination.

Our contributions are summarized as follows:

\paragraph{\textbf{A Benchmark for Systematic Evaluation.}}
We construct the first benchmark specifically designed for variant contamination detection, covering two representative domains: mathematical reasoning and general reasoning. Through controlled variant generation and filtering, this benchmark enables rigorous, reproducible evaluation of contamination detection methods across different models, scales, and target domains.

\paragraph{\textbf{A Novel Detection Framework.}}
We propose DVD, a training-free variant contamination detection method that relies solely on model generation behavior. By analyzing the variance of output distributions across multiple stochastic decoding runs, DVD captures anomalous fluctuations exhibited by models on contaminated queries, effectively penetrating surface-level paraphrasing to detect variant contamination. Experiments show that DVD significantly outperforms existing methods across models, scales, and domains, while remaining highly robust to decoding hyperparameters.

\section{Related Work}

Existing approaches for data contamination detection can be broadly divided into two categories.  

\paragraph{Sampling and Output-Matching-Based Methods}  
This line of research primarily relies on the similarity between model generations and reference answers, or on detecting anomalous patterns within the output distribution. Representative works include reference-instance matching based on overlap measures \cite{golchin2023time}; the CDD method, which conducts multiple random samplings alongside one greedy decoding under the same prompt, and uses the edit distance between greedy and stochastic outputs to approximate the output distribution and detect sharp modes caused by memorization \cite{khandelwal2019generalization}; and the DCQ method, which compares model preferences between original inputs and their perturbed variants to identify contamination \cite{golchin2025data}. Moreover, membership inference has also been applied in this context, where the loss difference between a target sample and synthetic neighbors serves as an indicator of contamination \cite{mattern2023membership}. Overall, these methods are effective for detecting verbatim memorization, yet remain limited by their reliance on shallow surface-level measures.

\paragraph{Perplexity-Based Methods}  
In contrast to sampling-and-matching-based approaches, another class of methods focuses on detecting contamination through the abnormally high confidence that models assign to seen samples. For example, the MIN-K\% PROB method examines the average log-likelihood of low-probability tokens to determine whether a sample appears in the training set \cite{shi2023detecting}. Similarly, \cite{oren2023proving} demonstrates that a model’s ability to recall the order of training samples itself constitutes strong evidence of data leakage. Compared to the former category, perplexity-based methods provide a more direct quantification of model bias toward training data. However, their effectiveness is likewise constrained to verbatim memorization; once samples undergo semantic rewriting or structural perturbation, perplexity-level differences are often largely obscured, leading to a significant drop in detection performance.

\paragraph{Our Approach}  
Motivated by the limitations of the above methods, we propose the DVD approach, which overcomes the dependence on shallow similarity measures or overall perplexity levels. Although CDD also relies on multiple samplings to construct an output distribution, its core remains restricted to edit-distance-based comparisons, failing to capture the true probabilistic dynamics underlying text generation. In contrast, DVD employs temperature sampling to generate multiple responses and systematically analyzes the variance of low-probability tokens, defined as synthetic difficulty. The key insight is that contaminated samples alternate between a "memorization-dependent state" and a "perturbation-drift state“ resulting in substantially higher variance across generations. Uncontaminated samples, by contrast, remain consistently in the drift state, with variance reflecting only natural noise. By incorporating variance decomposition into a mixture-distribution framework, DVD fundamentally captures these deep probabilistic dynamics, thereby achieving superior performance in detecting semantic-variant contamination compared to existing methods.

\section{Variant Contamination}  
This section introduces the formal definition of the \textbf{Variant Contamination Detection (VCD)} task (\ref{td}) and describes in detail the construction of a benchmark dataset tailored for reliable variant contamination detection (\ref{bc}).

\subsection{Task Definition}
\label{td}

We define \textbf{variant contamination} as the scenario in which, during training, a model is exposed to samples that are logically equivalent to those in the test set but differ in surface form. Such variants may diverge in semantics, syntax, or narrative style, yet preserve the same underlying solution space, thereby allowing the model to perform as if it had previously observed the test instance.  

Formally, let $x$ denote a test instance and let $f$ be a semantic abstraction function that extracts the core informational content of $x$. A variant of $x$ is then rigorously defined as follows:
\begin{equation}
v = \tau(x), \quad \text{such that } f(v) = f(x),
\end{equation}  
where $\tau$ is a transformation preserving the core semantics of $x$. If such a variant $v$ appears in the training corpus of model $M$, we say that $M$ is contaminated on test instance $x$. Importantly, unlike exact duplicates, variants may differ substantially from $x$ in vocabulary, phrasing, or narrative structure, while remaining equivalent in required knowledge, logical dependencies, and trajectory.

The goal of the VCD task is thus to reliably identify, within a model’s test set, which instances $x$ have been subject to contamination by semantically equivalent variants present in training.

\subsection{Benchmark Construction}  
\label{bc}
Variant contamination commonly arises in the context of data augmentation. To systematically evaluate the extent of variant contamination in large language models (LLMs), we construct a dedicated benchmark dataset. The construction leverages mainstream data augmentation techniques~\cite{shorten2019survey,shorten2021text,maharana2022review} and incorporates two widely used benchmarks: Omni-Math~\cite{gao2024omni} and SuperGPQA~\cite{du2025supergpqa}. As illustrated in Figure~\ref{fig:pipeline}, we employ GPT-4o~\cite{hurst2024gpt} to produce semantically equivalent variants from the original problem.

\paragraph{Initial Question and Solution Verification}  
In the initial stage, we first verify the original problem–solution pair $(x, y)$.  
If the problem already comes with a standardized solution, it directly proceeds to the next step;  
Otherwise (e.g., in the SuperGPQA dataset), GPT-4o~\cite{hurst2024gpt} is employed to generate gold-standard answers.
This ensures each problem is paired with a reference solution, forming the foundation for subsequent variant generation.
Formally, given the training set:  
\begin{equation}
D = \{(x_i, y_i)\}_{i=1}^N,
\end{equation}  
we take $(x_i, y_i)$ as input in preparation for generating corresponding variants. 

\paragraph{Variant Generation}
In this stage, we adopt mainstream data augmentation techniques~\cite{shorten2019survey,shorten2021text,maharana2022review} to generate a set of semantically equivalent variants $(x_v, y_v)$ for each original problem (see Table~\ref{tab:variant_methods} and Figure~\ref{fig:Variant_prompt}).  
Specifically, we define a transformation set:  
\begin{equation}
T = \{T_{\text{ent}}, T_{\text{scn}}, T_{\text{num}}, T_{\text{nar}}\},
\end{equation}  
covering four categories: entity substitution, scenario conversion, numerical rewriting, and narrative restructuring.  
Through these surface-level transformations, we construct the variant set:  
\begin{equation}
\begin{array}{cc}
     V(x_i) = \{v_i^{(1)}, \dots, v_i^{(m)}\},  \\
     \text{where } f(v_i^{(j)}) = f(x_i).
\end{array}
\end{equation}  
To guarantee semantic equivalence and correctness, rejection sampling is applied during generation (see Figure~\ref{fig:Reject_prompt}), with GPT-4o providing high-quality candidate variant answers.

\paragraph{Variant Filter}
Finally, GPT-4o is employed as a filter to conduct quality control over the generated variants~\cite{liu2025augmenting}.  
The filtering procedure consists of two steps: first, checking whether the information remains unchanged; second, performing a quality evaluation of the solution.  
Only when both conditions are satisfied is the variant pair $(x_v, y_v)$ accepted.  
Ultimately, these high-quality variant samples are injected into the training set to simulate test contamination, enabling systematic evaluation of whether existing detection methods can identify variant-contaminated test instances.

\begin{table*}[t]  
\centering
\small
\caption{Variant generation strategies used to simulate contamination.}
\label{tab:variant_methods}
\begin{tabular}{p{0.3\textwidth} p{0.65\textwidth}}
\toprule
\textbf{Method} & \textbf{Description} \\
\midrule
Entity substitution & Replace referents, variable names, and object categories while maintaining consistency in type and context. \\
Scenario transformation & Alter the background setting and narrative context, while preserving logical dependencies and constraint structures. \\
Numerical rewriting & Resample parameters under solvability constraints and update derivations and intermediate values for consistency. \\
Narrative structure transformation & Rearrange syntax or rewrite step-by-step analysis into a paragraph-style narrative while preserving semantic meaning. \\
\bottomrule
\end{tabular}
\end{table*}

\section{Method}
This paper proposes a method named DVD (\textbf{D}etection via \textbf{V}ariance of generation \textbf{D}istribution) grounded in modeling the distribution of model outputs. The core idea is to generate multiple responses under a fixed prompt using temperature sampling, thereby capturing fluctuations in low-probability regions of the model’s output distribution. These fluctuations serve as key signals for detecting contamination. More specifically, when a test sample appears in the training set, the model may operate in two distinct generative states. The first is memory adherence, where generation is guided by memorized templates internalized during training. The second is perturbation drift, where generation is primarily driven by stochastic perturbations introduced by temperature sampling, leading to free-form exploratory outputs. Memory adherence reflects the model’s reliance on training-based recall, while perturbation drift captures the natural randomness of unconstrained generation. If a test sample is contaminated, the model alternates between these two states, producing substantial variability in the conditional likelihoods of low-probability tokens. In contrast, for uncontaminated samples, the absence of reliable memory templates constrains the model to remain in a perturbation drift state, where tail-token probabilities mainly reflect inherent noise and thus exhibit only minor fluctuations. Based on this observation, we design the variance of synthetic difficulty as the contamination detection criterion.

\subsection{Temperature Sampling}
For each test sample $x_i$, we apply temperature sampling at test time under a fixed prompt $p$ to generate $N$ candidate responses $\{a_i^{(1)},a_i^{(2)},\dots,a_i^{(N)}\}$. Each response is concatenated with the prompt to form the complete input:
\begin{equation}
s_i^{(k)} = (p, a_i^{(k)}).
\end{equation}

Temperature sampling introduces stochastic perturbations, enabling the collection of diverse outputs for the same test sample. In uncontaminated cases, generation consistently remains in a perturbation drift state, and temperature perturbations do not alter the statistics of low-probability tokens. In contaminated cases, however, generation alternates between memory adherence and perturbation drift. Temperature perturbations amplify the disparity between template-based and non-template-based responses, causing tail tokens to exhibit more pronounced fluctuations.

\subsection{Synthetic Difficulty Modeling}
To quantify such fluctuations, we define the notion of \textbf{synthetic difficulty}. For each sequence $s_i^{(k)}$, we select the $k$ least probable tokens in the response, compute the sum of their log-likelihoods, and normalize by sequence length $T_i^{(k)}$:
\begin{equation}
D_i^{(k)} = \frac{1}{T_i^{(k)}} \sum_{j=1}^{k} \log P_\theta\!\left(t^{(j)} \mid s_i^{(k)}\right).
\end{equation}

This statistic captures local uncertainty in the tail region of the distribution. Unlike global perplexity, tail-token probabilities are more sensitive to the presence of training-set memorization. If a test sample is contaminated, $D_i^{(k)}$ varies markedly across generations due to the alternation between memory adherence and perturbation drift. If uncontaminated, tail probabilities primarily reflect noise, yielding relatively stable values of $D_i^{(k)}$ across multiple generations.

Given the synthetic difficulty set $\{D_i^{(1)}, D_i^{(2)}, \dots, D_i^{(N)}\}$, we define the DVD indicator as their sample variance:
\begin{equation}
\begin{array}{cc}
     \mathrm{DVD}_i = \frac{1}{N} \sum_{k=1}^{N} \left(D_i^{(k)} - \overline{D}_i \right)^2, \\
     \overline{D}_i = \frac{1}{N} \sum_{j=1}^{N} D_i^{(j)}.
\end{array}
\end{equation}
This indicator characterizes the fluctuation of synthetic difficulty. According to the variance decomposition principle for mixture distributions, if a test sample is contaminated, the distribution of synthetic difficulty can be regarded as a mixture of memory states and drift states, which differ in expectation, thereby inflating the overall variance. If uncontaminated, synthetic difficulty arises from a single state, and variance remains low.

More specifically, contaminated samples can be modeled as a mixture of two latent generation states: the memory-adhering state ($Z=M$) dominated by training memorization, and the unconstrained perturbation-drift state ($Z=U$). Let $\pi_{M} = \Pr(Z = M)$, $\pi_{U} = \Pr(Z = U)$, with $\pi_M + \pi_U = 1$. Then,
\begin{equation}
\mu = \pi_M \mu_M + \pi_U \mu_U ,
\end{equation}
\begin{equation}
\begin{array}{cc}
     \mathrm{Var}(X) = 
     \pi_M \left( \sigma_M^{2} + (\mu_M - \mu)^{2} \right) + \\
     \pi_U \left( \sigma_U^{2} + (\mu_U - \mu)^{2} \right).
\end{array}
\end{equation}

Here, $\mu_M$ and $\mu_U$ denote the expectations under the memory and drift states, respectively. Since the memory state relies on templates encountered during training, its synthetic difficulty is generally lower than that of the drift state, i.e., $\mu_M > \mu_U$ empirically. By the decomposition of within-group and between-group variance, if the two states differ substantially in expectation, the overall variance of the mixture will exceed that of a single distribution. This theoretical grounding demonstrates the effectiveness of our method in distinguishing contaminated from uncontaminated samples.

\section{Experiments}
In this section, we simulate a variant-contamination scenario based on the constructed variant dataset (see Section~\ref{bc}) and perform a comprehensive evaluation of our method against a range of baseline approaches under this setting.We exclude closed-source models from our study because they are not practically trainable/fine-tunable, making it difficult to simulate the variant-contamination setting. Detailed experimental configurations are provided in Section~\ref{sec:experimental_setup}, large language model fine-tuning details are described in Section~\ref{sec:Fine-tuning-details}, and the experimental results are reported in Section~\ref{rs}.

\subsection{Experimental Setup}

\label{sec:experimental_setup}

\paragraph{Model selection:} To comprehensively assess the robustness of our variant-contamination detection method, we compare models along multiple dimensions: parameter scale (Qwen2.5-1.5B-Instruct \cite{qwen2.5}, Qwen2.5-3B-Instruct \cite{qwen2.5}, Qwen2.5-7B-Instruct \cite{qwen2.5}, Qwen2.5-32B-Instruct \cite{qwen2.5}), architecture (Qwen2.5 vs.\ Llama3.1 \cite{dubey2024llama}), and fine-tuning strategy (full-parameter fine-tuning vs.\ LoRA fine-tuning).

\paragraph{Baselines:} To validate the effectiveness of our method, we compare it with the following baselines: 1)\textit{Embedding Similarity} \cite{dong2024generalization}: computes the similarity between answers using embeddings produced by the base model; 2)\textit{Perplexity} \cite{li2023open}: computes the perplexity of the original answer given the prompt; 3)\textit{Min-k\% Probability} \cite{shi2023detecting}: computes the average probability over the lowest k\% token probabilities of the original answer given the prompt; 4)\textit{Min-k\%++ Probability} \cite{zhang2025min}: an enhanced variant of Min-k\%, which normalizes and calibrates token log-probabilities using statistics (mean and standard deviation) of the class distribution over the model vocabulary, and takes the average over the lowest k\% calibrated scores as the detection score; 5)\textit{CDD} \cite{dong2024generalization}: measures the sharpness of the output distribution via edit distance; 6)\textit{Zlib} \cite{zhang2025min}: computes the Zlib compression entropy of the original answer given the prompt; 7)\textit{Loss} \cite{zhang2025min}: computes the loss of the original answer given the prompt. The hyperparameters specific to our method were set as follows: the number of minimum-probability tokens \textit{k} was fixed at 20, and the number of samples \textit{N} was set to 50.

\begin{table*}[h]
\centering
\small
\setlength{\tabcolsep}{3pt}
\begin{tabular}{l *{4}{c} *{4}{c}}
\toprule
\textbf{Method} & \multicolumn{4}{c}{\textbf{Omni-MATH}} & \multicolumn{4}{c}{\textbf{SuperGPQA}} \\
\cmidrule(lr){2-5} \cmidrule(l){6-9}
& \textbf{Qwen1.5B} & \textbf{Qwen3B} & \textbf{Qwen7B} & \textbf{Qwen32B}
& \textbf{Qwen1.5B} & \textbf{Qwen3B} & \textbf{Qwen7B} & \textbf{Qwen32B} \\
\midrule
Min-K\%++ & \cellcolor{pink!20}0.694 & \cellcolor{pink!20}0.693 & \cellcolor{pink!20}0.680 & \cellcolor{pink!20}0.681 
          & 0.422 & 0.463 & 0.435 & 0.436 \\
CDD & 0.494 & 0.495 & 0.512 & 0.507 
    & 0.496 & 0.504 & 0.518 & 0.503 \\
Min-K & 0.538 & 0.560 & 0.572 & 0.578 
      & 0.501 & 0.492 & 0.511 & \cellcolor{pink!20}0.539 \\
Perplexity & 0.544 & 0.549 & 0.557 & 0.556 
           & 0.517 & 0.520 & 0.513 & 0.517 \\
Loss & 0.626 & 0.637 & 0.650 & 0.635 
     & 0.404 & 0.406 & 0.408 & 0.409 \\
Zlib & 0.573 & 0.576 & 0.583 & 0.581 
     & 0.425 & 0.430 & 0.428 & 0.427 \\
\makecell[l]{EM} & 0.521 & 0.506 & 0.533 & 0.505 
                                     & \cellcolor{pink!20}0.531 & \cellcolor{pink!20}0.529 & \cellcolor{pink!20}0.524 & 0.521 \\
DVD (Ours) & \cellcolor{orange!20}0.744 & \cellcolor{orange!20}0.747 & \cellcolor{orange!20}0.734 & \cellcolor{orange!20}0.667 
           & \cellcolor{orange!20}0.770 & \cellcolor{orange!20}0.708 & \cellcolor{orange!20}0.740 & \cellcolor{orange!20}0.743 \\
\bottomrule
\end{tabular}
\caption{Performance comparison of different detection methods on Omni-MATH and SuperGPQA datasets with full Fine-tuning (1 epoch). EM denotes the Embedding-similarity method. The notation “QwenXB” in the table refers to the Qwen2.5-XB-Instruct model, where X denotes the model's parameter count}
\label{tab:combined_performance}
\end{table*}

\begin{table*}[h]
\centering
\small
\setlength{\tabcolsep}{3pt}
\begin{tabular}{l *{4}{c} *{4}{c}}
\toprule
\textbf{Method} & \multicolumn{4}{c}{\textbf{Omni-MATH}} & \multicolumn{4}{c}{\textbf{SuperGPQA}} \\
\cmidrule(lr){2-5} \cmidrule(l){6-9}
& \textbf{Qwen1.5B} & \textbf{Qwen3B} & \textbf{Qwen7B} & \textbf{Qwen32B}
& \textbf{Qwen1.5B} & \textbf{Qwen3B} & \textbf{Qwen7B} & \textbf{Qwen32B} \\
\midrule
Min-K\%++ & \cellcolor{pink!20}0.646 & \cellcolor{pink!20}0.621 & \cellcolor{pink!20}0.648 & \cellcolor{pink!20}0.608 
          & 0.415 & 0.456 & 0.364 & 0.369 \\
CDD & 0.501 & 0.503 & 0.513 & 0.507 
    & 0.520 & 0.517 & 0.584 & 0.600 \\
Min-K & 0.549 & 0.505 & 0.531 & 0.536 
      & 0.497 & 0.519 & 0.414 & 0.341 \\
Perplexity & 0.572 & 0.543 & 0.567 & 0.557 
           & 0.478 & 0.495 & 0.404 & 0.370 \\
Loss & 0.572 & 0.556 & 0.567 & 0.563 
     & 0.379 & 0.372 & 0.334 & 0.333 \\
Zlib & 0.549 & 0.542 & 0.550 & 0.550 
     & 0.409 & 0.400 & 0.378 & 0.376 \\
\makecell[l]{EM} & 0.590 & 0.608 & 0.544 & 0.563 
                                     & \cellcolor{pink!20}0.580 & \cellcolor{pink!20}0.594 & \cellcolor{pink!20}0.599 & \cellcolor{pink!20}0.712 \\
DVD (Ours) & \cellcolor{orange!20}0.771 & \cellcolor{orange!20}0.745 & \cellcolor{orange!20}0.731 & \cellcolor{orange!20}0.715 
           & \cellcolor{orange!20}0.674 & \cellcolor{orange!20}0.737 & \cellcolor{orange!20}0.700 & \cellcolor{orange!20}0.751 \\
\bottomrule
\end{tabular}
\caption{Performance comparison of detection methods on Omni-MATH and SuperGPQA with Lora Fine-tuning (10 epochs). EM denotes the Embedding-similarity method. The notation “QwenXB” in the table refers to the Qwen2.5-XB-Instruct model, where X denotes the model's parameter count}
\label{tab:lora_10epoch_combined}
\end{table*}


\subsection{Experimental Results}
\label{rs}

We evaluate the performance of our proposed method, \textbf{DVD}, against several baselines on two distinct datasets: \textit{Omni-MATH} and \textit{SuperGPQA}. The results, measured by AUC, are summarized in Tables~\ref{tab:combined_performance} and \ref{tab:lora_10epoch_combined}. Across all settings, DVD consistently and significantly outperforms all baseline approaches, demonstrating superior detection accuracy and cross-domain robustness.

\paragraph{Superiority Over Log-Probability and Loss-Based Baselines:} Traditional detection methods such as \textit{Loss}, \textit{Perplexity}, and \textit{Zlib} exhibit highly unstable performance across different scenarios. For instance, while the \textit{Loss} method achieves moderate results on Omni-MATH (e.g., AUC of 0.626 to 0.664 in Table~\ref{tab:combined_performance}), its performance deteriorates sharply on the SuperGPQA dataset, with AUC values significantly below the random-chance threshold of 0.5 (e.g., 0.333 to 0.409 in Table~\ref{tab:lora_10epoch_combined}). Importantly, an AUC below 0.5 indicates that the model’s predictions are systematically inverted relative to the true labels. Specifically, samples with higher loss are more likely to be incorrectly classified as clean, while those with lower loss are misidentified as contaminated. Similarly, \textit{Perplexity} and \textit{Zlib} yield AUCs consistently in the range of 0.4 to 0.5 on SuperGPQA, reflecting a similar tendency to produce judgments that contradict the actual contamination status. This clearly demonstrates that simple likelihood- or compression-based metrics are highly sensitive to data distribution and training configurations, lacking not only robustness but also the basic reliability required for effective detection of variant contamination.

\paragraph{Analysis of Min-K and Min-K\%++:} The \textit{Min-K} and \textit{Min-K\%++} methods, which focus on the likelihood of the least probable tokens, show a specialized but fragile advantage. \textit{Min-K\%++} is the strongest baseline on the Omni-MATH dataset, achieving AUCs between 0.608 and 0.694. However, its effectiveness diminishes on SuperGPQA, where it drops as low as 0.278 (Table~\ref{tab:lora_10epoch_combined}). This indicates that while focusing on outlier token probabilities helps in structured mathematical domains, it fails to generalize to complex, semantic-heavy reasoning tasks where the "variant" nature of the data is not captured by local token statistics.

\paragraph{Comparison with Distributional and Similarity Measures:} We also compared our method against \textit{CDD} (based on edit distance) and \textit{Embedding-similarity}. \textit{CDD} consistently performs near the level of random guessing (AUC $\approx$ 0.50) on Omni-MATH, suggesting that surface-level text fluctuations are insufficient for detection in diverse mathematical contexts. On SuperGPQA, \textit{CDD} improves slightly (up to 0.600) but remains uncompetitive. \textit{Embedding-similarity} proves to be a more robust baseline, particularly on SuperGPQA where it achieves AUCs between 0.521 and 0.712. Nevertheless, it still lags behind DVD by a significant margin. This confirms that while semantic similarity captures some distributional shifts, it cannot match the discriminative power of DVD’s difficulty-fluctuation modeling.

\paragraph{Effectiveness Under Different Fine-tuning Regimes:} The experimental results across Full Fine-tuning (1 epoch) and LoRA (10 epochs) highlight DVD's versatility. In the challenging 10-epoch LoRA setting on Omni-MATH (Table~\ref{tab:lora_10epoch_combined}), DVD achieves an AUC of 0.771 on Qwen-1.5B, while the next best baseline (\textit{Min-K\%++}) only reaches 0.646. Even when the model is heavily fine-tuned, DVD effectively captures the "synthetic difficulty" signatures that distinguish contaminated variants from clean data.

\subsection{Ablation Study}
\begin{figure}[ht]
    \centering
    \includegraphics[width=\columnwidth]{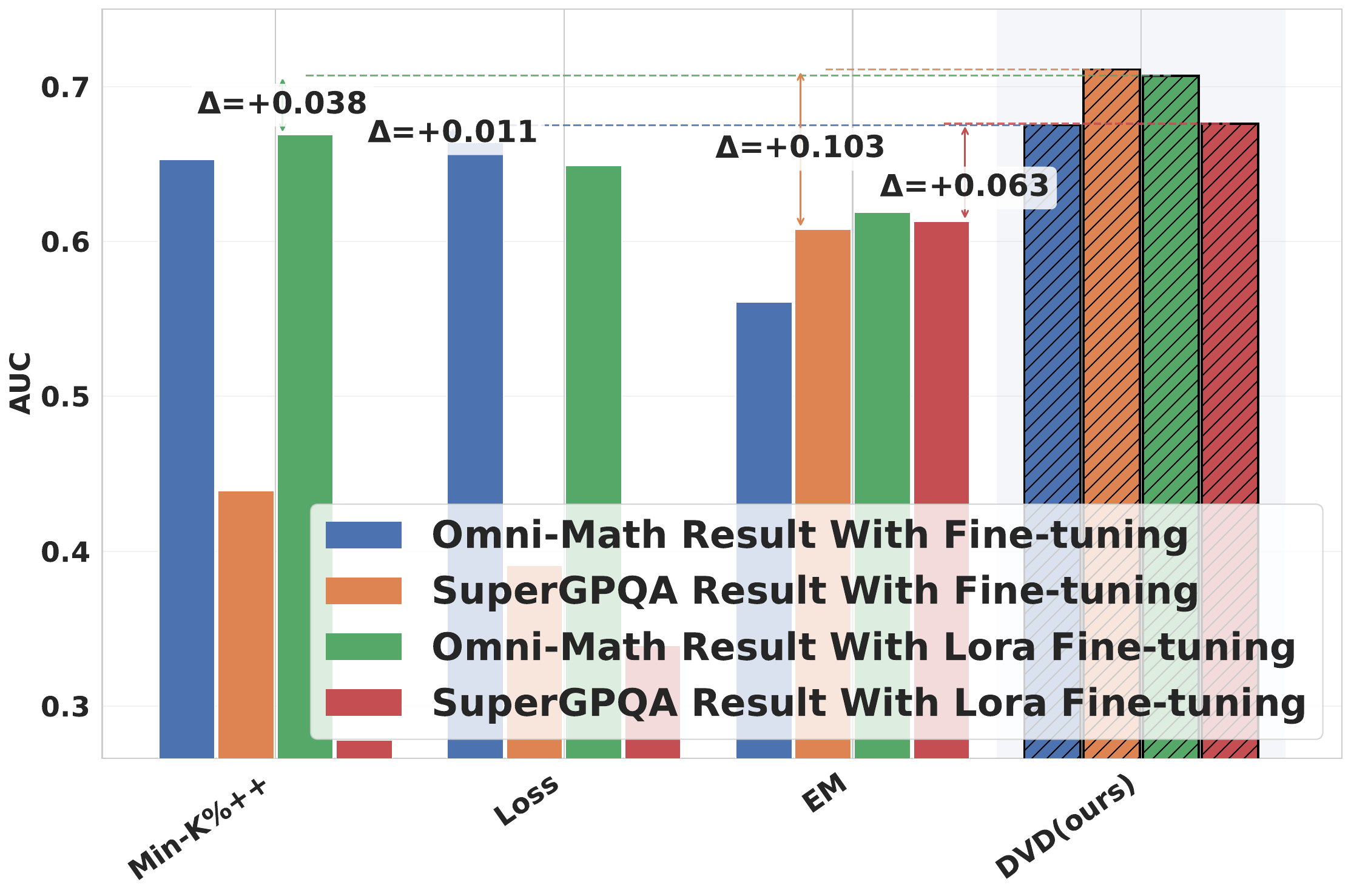}
    \caption{Performance comparison of the DVD method and other baselines on the Llama architecture (using the Llama3.1-8B-Instruct model). EM denotes the Embedding-similarity method. A complete visualization of the results is provided in Figure~\ref{fig:ablationStudyall} of Appendix~\ref{sec:ablationStudyall}.}
    \label{fig:ablationStudy}
\end{figure}
\vspace{-6pt}
\paragraph{Robustness Across Model Scales and Architectures:} Our method exhibits remarkable stability across different model sizes (from Qwen-1.5B to 32B) and architectures (Qwen and Llama). As shown in Table~\ref{tab:combined_performance}, DVD maintains high AUCs (0.667--0.747) regardless of the parameter count. Notably, on the Llama-8B model, DVD consistently provides a substantial gain over the best baselines (shown in Figure \ref{fig:ablationStudy}). While other methods like \textit{Min-K\%++} or \textit{Embedding-similarity} fluctuate wildly depending on the model scale, DVD’s performance remains consistently high, validating its architecture-agnostic nature.
\begin{figure}[ht]
    \centering
    \includegraphics[width=\columnwidth]{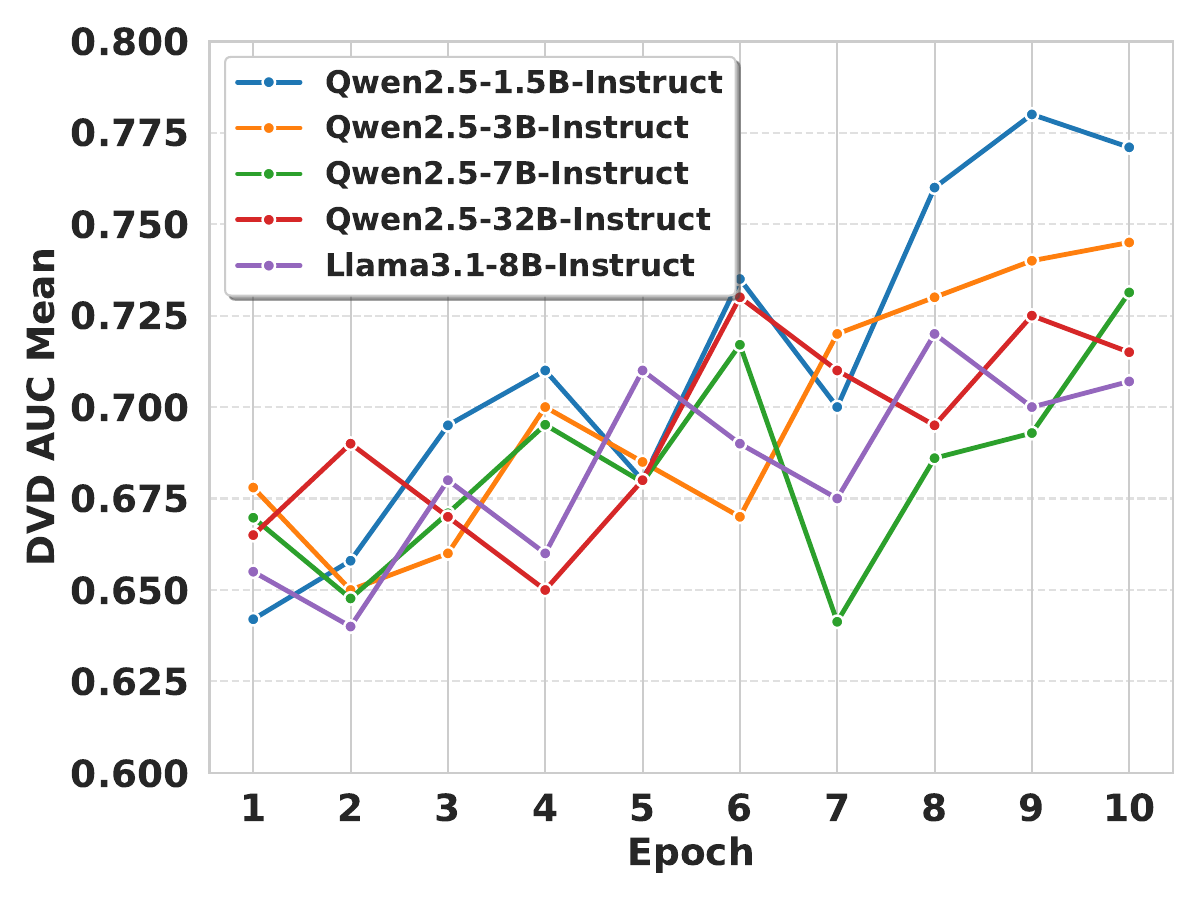}
    \caption{Average performance of the DVD method across different epochs.}
    \label{fig:ablationStudy}
\end{figure}
\vspace{-6pt}
\paragraph{Robustness of DVD Across Training Epochs and Model Scales:}
As shown in Figure \ref{fig:ablationStudy}, each data point represents the average result over ten independent runs to mitigate the influence of randomness. DVD demonstrates consistently high robustness across different training epochs. When evaluated on various instruction tuned large language models, including the Qwen2.5 series and Llama3.1-8BInstruct, DVD achieves strong AUC scores for detecting model variant contamination as early as the initial training stages. Moreover, detection performance generally improves with increasing model scale; for instance, Qwen2.5 32B Instruct attains a higher and more stable AUC. Collectively, these results indicate that DVD exhibits consistent and reliable detection capability across models of varying sizes and architectures, as well as across different training epochs.

\section{Conclusion}

This work systematically uncovers the overlooked problem of variant contamination in large language models, establishes the first benchmark dedicated to this issue, and proposes DVD as a principled solution. DVD effectively identifies contaminated samples by modeling fluctuations in synthesis difficulty across multiple generations, significantly outperforming conventional approaches based on log probability, distributional properties, and similarity metrics. Evaluated on the proposed benchmark, which encompasses Omni-MATH and SuperGPQA, DVD demonstrates consistently high accuracy and strong cross-domain robustness across diverse training configurations and model scales, offering a reliable tool to mitigate contamination risks and enable fairer, more trustworthy evaluation of large language models.

\section{Limitation}
For tasks with open-ended answers, underspecified problem statements, or multiple valid reasoning paths, temperature sampling naturally induces greater output diversity, thereby elevating the baseline level of variance in the generation distribution. As a result, when comparing DVD scores across tasks or domains, it is generally necessary to adopt unified prompt templates, length constraints, and decoding configurations, and to apply task- or category-specific calibration or relative scoring schemes  to ensure interpretability and comparability. Moreover, because DVD derives its signal from local uncertainty fluctuations in low-probability tokens, it is sensitive to generation length, stopping criteria, and answer formats, all of which can alter the composition and proportion of tail tokens and thus affect the stability of synthetic difficulty estimation. At a practical level, DVD constructs a local generation distribution via repeated sampling, which necessitates balancing the number of samples against detection stability in environments with limited query budgets or restricted access. Finally, the benchmark construction and contamination simulation in this work rely primarily on LLM-generated and LLM-filtered semantic variants injected through fine-tuning; while this setup is reproducible and well controlled, variant contamination in real pretraining corpora may arise from more diverse sources—such as cross-domain reuse, templated rewriting, heterogeneous annotation styles, or retrieval-augmented pipelines—whose behavioral signatures and decision boundaries warrant further systematic characterization in future work.

\bibliography{custom}

@article{li2023open,
  title={An open source data contamination report for large language models},
  author={Li, Yucheng and Guerin, Frank and Lin, Chenghua},
  journal={arXiv preprint arXiv:2310.17589},
  year={2023}
}

@article{cheng2025survey,
  title={A survey on data contamination for large language models},
  author={Cheng, Yuxing and Chang, Yi and Wu, Yuan},
  journal={arXiv preprint arXiv:2502.14425},
  year={2025}
}

@article{deng2023investigating,
  title={Investigating data contamination in modern benchmarks for large language models},
  author={Deng, Chunyuan and Zhao, Yilun and Tang, Xiangru and Gerstein, Mark and Cohan, Arman},
  journal={arXiv preprint arXiv:2311.09783},
  year={2023}
}

@article{balloccu2024leak,
  title={Leak, cheat, repeat: Data contamination and evaluation malpractices in closed-source LLMs},
  author={Balloccu, Simone and Schmidtov{\'a}, Patr{\'\i}cia and Lango, Mateusz and Du{\v{s}}ek, Ond{\v{r}}ej},
  journal={arXiv preprint arXiv:2402.03927},
  year={2024}
}

@article{chang2024survey,
  title={A survey on evaluation of large language models},
  author={Chang, Yupeng and Wang, Xu and Wang, Jindong and Wu, Yuan and Yang, Linyi and Zhu, Kaijie and Chen, Hao and Yi, Xiaoyuan and Wang, Cunxiang and Wang, Yidong and others},
  journal={ACM transactions on intelligent systems and technology},
  volume={15},
  number={3},
  pages={1--45},
  year={2024},
  publisher={ACM New York, NY}
}

@article{xu2024benchmarking,
  title={Benchmarking benchmark leakage in large language models},
  author={Xu, Ruijie and Wang, Zengzhi and Fan, Run-Ze and Liu, Pengfei},
  journal={arXiv preprint arXiv:2404.18824},
  year={2024}
}

@article{gpt3,
  title={Language models are few-shot learners},
  author={Brown, Tom and Mann, Benjamin and Ryder, Nick and Subbiah, Melanie and Kaplan, Jared D and Dhariwal, Prafulla and Neelakantan, Arvind and Shyam, Pranav and Sastry, Girish and Askell, Amanda and others},
  journal={Advances in neural information processing systems},
  volume={33},
  pages={1877--1901},
  year={2020}
}

@article{palm,
  title={Palm: Scaling language modeling with pathways},
  author={Chowdhery, Aakanksha and Narang, Sharan and Devlin, Jacob and Bosma, Maarten and Mishra, Gaurav and Roberts, Adam and Barham, Paul and Chung, Hyung Won and Sutton, Charles and Gehrmann, Sebastian and others},
  journal={Journal of Machine Learning Research},
  volume={24},
  number={240},
  pages={1--113},
  year={2023}
}

@article{gpt4,
  title={Gpt-4 technical report},
  author={Achiam, Josh and Adler, Steven and Agarwal, Sandhini and Ahmad, Lama and Akkaya, Ilge and Aleman, Florencia Leoni and Almeida, Diogo and Altenschmidt, Janko and Altman, Sam and Anadkat, Shyamal and others},
  journal={arXiv preprint arXiv:2303.08774},
  year={2023}
}

@article{llama,
  title={Llama: Open and efficient foundation language models},
  author={Touvron, Hugo and Lavril, Thibaut and Izacard, Gautier and Martinet, Xavier and Lachaux, Marie-Anne and Lacroix, Timoth{\'e}e and Rozi{\`e}re, Baptiste and Goyal, Naman and Hambro, Eric and Azhar, Faisal and others},
  journal={arXiv preprint arXiv:2302.13971},
  year={2023}
}

@article{gemini,
  title={Gemini: A family of highly capable multimodal models, 2024},
  author={Team, Gemini and Anil, R and Borgeaud, S and Wu, Y and Alayrac, JB and Yu, J and Soricut, R and Schalkwyk, J and Dai, AM and Hauth, A and others},
  journal={arXiv preprint arXiv:2312.11805},
  volume={10},
  year={2024}
}

@article{sainz2023nlp,
  title={NLP evaluation in trouble: On the need to measure LLM data contamination for each benchmark},
  author={Sainz, Oscar and Campos, Jon Ander and Garc{\'\i}a-Ferrero, Iker and Etxaniz, Julen and de Lacalle, Oier Lopez and Agirre, Eneko},
  journal={arXiv preprint arXiv:2310.18018},
  year={2023}
}

@article{golchin2023time,
  title={Time travel in llms: Tracing data contamination in large language models},
  author={Golchin, Shahriar and Surdeanu, Mihai},
  journal={arXiv preprint arXiv:2308.08493},
  year={2023}
}

@article{khandelwal2019generalization,
  title={Generalization through memorization: Nearest neighbor language models},
  author={Khandelwal, Urvashi and Levy, Omer and Jurafsky, Dan and Zettlemoyer, Luke and Lewis, Mike},
  journal={arXiv preprint arXiv:1911.00172},
  year={2019}
}

@article{golchin2025data,
  title={Data contamination quiz: A tool to detect and estimate contamination in large language models},
  author={Golchin, Shahriar and Surdeanu, Mihai},
  journal={Transactions of the Association for Computational Linguistics},
  volume={13},
  pages={809--830},
  year={2025},
  publisher={MIT Press 255 Main Street, 9th Floor, Cambridge, Massachusetts 02142, USA~…}
}

@article{mattern2023membership,
  title={Membership inference attacks against language models via neighbourhood comparison},
  author={Mattern, Justus and Mireshghallah, Fatemehsadat and Jin, Zhijing and Sch{\"o}lkopf, Bernhard and Sachan, Mrinmaya and Berg-Kirkpatrick, Taylor},
  journal={arXiv preprint arXiv:2305.18462},
  year={2023}
}

@article{shi2023detecting,
  title={Detecting pretraining data from large language models},
  author={Shi, Weijia and Ajith, Anirudh and Xia, Mengzhou and Huang, Yangsibo and Liu, Daogao and Blevins, Terra and Chen, Danqi and Zettlemoyer, Luke},
  journal={arXiv preprint arXiv:2310.16789},
  year={2023}
}

@inproceedings{oren2023proving,
  title={Proving test set contamination in black-box language models},
  author={Oren, Yonatan and Meister, Nicole and Chatterji, Niladri S and Ladhak, Faisal and Hashimoto, Tatsunori},
  booktitle={The Twelfth International Conference on Learning Representations},
  year={2023}
}

@article{dong2024generalization,
  title={Generalization or memorization: Data contamination and trustworthy evaluation for large language models},
  author={Dong, Yihong and Jiang, Xue and Liu, Huanyu and Jin, Zhi and Gu, Bin and Yang, Mengfei and Li, Ge},
  journal={arXiv preprint arXiv:2402.15938},
  year={2024}
}

@article{shorten2021text,
  title={Text data augmentation for deep learning},
  author={Shorten, Connor and Khoshgoftaar, Taghi M and Furht, Borko},
  journal={Journal of big Data},
  volume={8},
  number={1},
  pages={101},
  year={2021},
  publisher={Springer}
}

@article{maharana2022review,
  title={A review: Data pre-processing and data augmentation techniques},
  author={Maharana, Kiran and Mondal, Surajit and Nemade, Bhushankumar},
  journal={Global Transitions Proceedings},
  volume={3},
  number={1},
  pages={91--99},
  year={2022},
  publisher={Elsevier}
}

@article{shorten2019survey,
  title={A survey on image data augmentation for deep learning},
  author={Shorten, Connor and Khoshgoftaar, Taghi M},
  journal={Journal of big data},
  volume={6},
  number={1},
  pages={1--48},
  year={2019},
  publisher={Springer}
}

@article{gao2024omni,
  title={Omni-math: A universal olympiad level mathematic benchmark for large language models},
  author={Gao, Bofei and Song, Feifan and Yang, Zhe and Cai, Zefan and Miao, Yibo and Dong, Qingxiu and Li, Lei and Ma, Chenghao and Chen, Liang and Xu, Runxin and others},
  journal={arXiv preprint arXiv:2410.07985},
  year={2024}
}

@article{du2025supergpqa,
  title={Supergpqa: Scaling llm evaluation across 285 graduate disciplines},
  author={Du, Xinrun and Yao, Yifan and Ma, Kaijing and Wang, Bingli and Zheng, Tianyu and Zhu, King and Liu, Minghao and Liang, Yiming and Jin, Xiaolong and Wei, Zhenlin and others},
  journal={arXiv preprint arXiv:2502.14739},
  year={2025}
}

@article{hurst2024gpt,
  title={Gpt-4o system card},
  author={Hurst, Aaron and Lerer, Adam and Goucher, Adam P and Perelman, Adam and Ramesh, Aditya and Clark, Aidan and Ostrow, AJ and Welihinda, Akila and Hayes, Alan and Radford, Alec and others},
  journal={arXiv preprint arXiv:2410.21276},
  year={2024}
}

@inproceedings{liu2025augmenting,
  title={Augmenting math word problems via iterative question composing},
  author={Liu, Haoxiong and Zhang, Yifan and Luo, Yifan and Yao, Andrew C},
  booktitle={Proceedings of the AAAI Conference on Artificial Intelligence},
  volume={39},
  number={23},
  pages={24605--24613},
  year={2025}
}

@misc{qwen2.5,
    title = {Qwen2.5: A Party of Foundation Models},
    url = {https://qwenlm.github.io/blog/qwen2.5/},
    author = {Qwen Team},
    month = {September},
    year = {2024}
}

@article{dubey2024llama,
  title={The llama 3 herd of models},
  author={Dubey, Abhimanyu and Jauhri, Abhinav and Pandey, Abhinav and Kadian, Abhishek and Al-Dahle, Ahmad and Letman, Aiesha and Mathur, Akhil and Schelten, Alan and Yang, Amy and Fan, Angela and others},
  journal={arXiv e-prints},
  pages={arXiv--2407},
  year={2024}
}

@inproceedings{zhang2025min,
 author = {Zhang, Jingyang and Sun, Jingwei and Yeats, Eric and Ouyang, Yang and Kuo, Martin and Zhang, Jianyi and Yang, Hao and Li, Hai},
 booktitle = {International Conference on Representation Learning},
 editor = {Y. Yue and A. Garg and N. Peng and F. Sha and R. Yu},
 pages = {64845--64862},
 title = {Min-K\%++: Improved Baseline for Pre-Training Data Detection from Large Language Models},
 url = {https://proceedings.iclr.cc/paper_files/paper/2025/file/a2e3b4132ab2e0b7a21e6e75da7f91a9-Paper-Conference.pdf},
 volume = {2025},
 year = {2025}
}

@article{lunardi2025robustness,
  title={On Robustness and Reliability of Benchmark-Based Evaluation of LLMs},
  author={Lunardi, Riccardo and Della Mea, Vincenzo and Mizzaro, Stefano and Roitero, Kevin},
  journal={arXiv preprint arXiv:2509.04013},
  year={2025}
}

@article{sclar2023quantifying,
  title={Quantifying Language Models' Sensitivity to Spurious Features in Prompt Design or: How I learned to start worrying about prompt formatting},
  author={Sclar, Melanie and Choi, Yejin and Tsvetkov, Yulia and Suhr, Alane},
  journal={arXiv preprint arXiv:2310.11324},
  year={2023}
}

@article{zhao2024improving,
  title={Improving the robustness of large language models via consistency alignment},
  author={Zhao, Yukun and Yan, Lingyong and Sun, Weiwei and Xing, Guoliang and Wang, Shuaiqiang and Meng, Chong and Cheng, Zhicong and Ren, Zhaochun and Yin, Dawei},
  journal={arXiv preprint arXiv:2403.14221},
  year={2024}
}

\newpage

\appendix
\section{Appendix}
\subsection{Fine-tuning -details}
\label{sec:Fine-tuning-details}
To emulate variant contamination, we fine-tune the above models on our constructed variant-contamination dataset. To model resource-constrained scenarios, we adopt LoRA for parameter-efficient adaptation; this training is conducted on a single NVIDIA A800 GPU with the following settings: LoRA rank 8; Adam optimizer; 10 training epochs; initial learning rate $1e−5$; a cosine learning-rate scheduler with a warmup ratio of 0.1; per-GPU batch size 2; gradient accumulation steps 1; and bfloat16 precision. To model quality-prioritized scenarios, we additionally perform full-parameter fine-tuning on two NVIDIA A800 GPUs, using: Adam optimizer; 1 training epoch; initial learning rate $1e−5$; a cosine learning-rate scheduler with a warmup ratio of 0.1; per-GPU batch size 2; gradient accumulation steps 1; and bfloat16 precision
\subsection{Hyperparameter Sensitivity Analysis}

\begin{figure*}[ht]
    \centering
    \includegraphics[width=\textwidth]{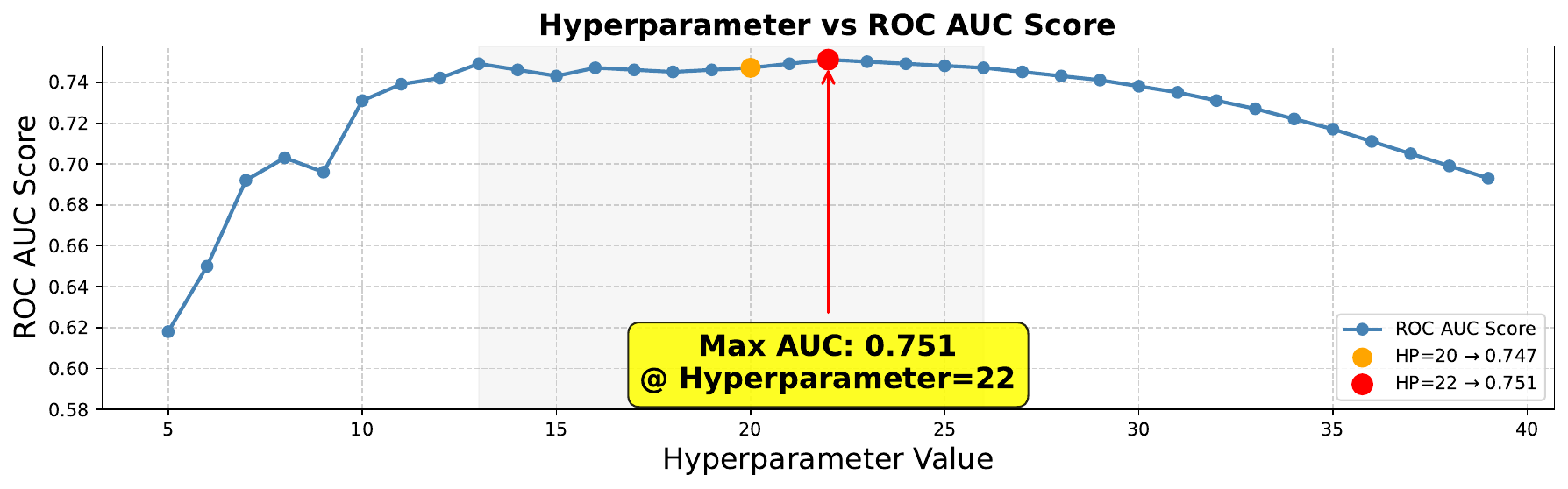}
    \caption{The DVD method demonstrates remarkable robustness across a wide range of hyperparameters.}
    \label{fig:hyperparameters_robustness}
\end{figure*}

To evaluate the sensitivity of the proposed DVD method to the key hyperparameter $M$ (i.e., the minimum number of low-probability tokens considered when calculating the synthetic difficulty), we conducted extensive experiments on the Qwen2.5-3B-Instruct model and the Omni-MATH variant dataset. The experimental results (Figure~\ref{fig:hyperparameters_robustness}) reveal both the effectiveness and moderate sensitivity of the method to $M$.

\textbf{Strong Performance with a Clear Optimal Region}  
The detection performance of the DVD method, measured by ROC AUC, peaks at \textbf{0.751} when $M = 22$. Notably, even at nearby values—such as $M = 20$ (AUC = 0.747)—performance remains high, indicating a \textbf{well-defined and broad performance peak} rather than extreme fragility. Across the full tested range (approximately $M = 5$ to $M = 35$), AUC scores stay consistently above 0.58 and reach a maximum of 0.751, demonstrating that the method is effective over a wide hyperparameter regime.

\textbf{Presence of a High-Performance Interval}  
Although the curve is not completely flat, a \textbf{robust high-performance interval} exists around $M \in [18, 26]$, where AUC remains above 0.74. This suggests that while fine-tuning $M$ can yield marginal gains, users can still achieve near-optimal detection performance by selecting $M$ within this practical interval—avoiding the need for exhaustive search while maintaining strong results.

\textbf{Clear Advantage Over Baseline Methods}  
Critically, even the \textbf{lower end} of the observed AUC range (e.g., $\sim$0.58 at extreme $M$ values) is comparable to or exceeds the performance of baseline methods such as Min-K\%++ Prob (0.693), Perplexity (0.549), and CDD (0.495). More importantly, the \textbf{peak performance (0.751)} substantially outperforms all baselines, confirming that the DVD method’s superiority is both significant and realizable with reasonable hyperparameter choices.

\textbf{Theoretical Interpretation}  
The unimodal shape of the AUC curve aligns with the underlying mechanism of DVD:  
\begin{itemize}
    \item When $M$ is too small (e.g., $M < 15$), the synthetic difficulty is estimated from too few tokens, leading to high variance and unreliable detection signals.
    \item When $M$ is too large (e.g., $M > 30$), the inclusion of medium-probability tokens dilutes the signal from truly ``difficult'' (low-probability) tokens, slightly degrading discriminative power.
    \item Around $M = 22$, the method strikes an optimal balance—capturing enough low-probability tokens for stable variance estimation while avoiding noise from less informative tokens.
\end{itemize}
This behavior reflects a \textbf{principled trade-off} inherent in the design of DVD, rather than arbitrary sensitivity. The existence of a clear, high-performing region further supports the method’s practical utility.

\subsection{Case Study}
\begin{figure*}[ht]
    \centering
    \includegraphics[width=\textwidth]{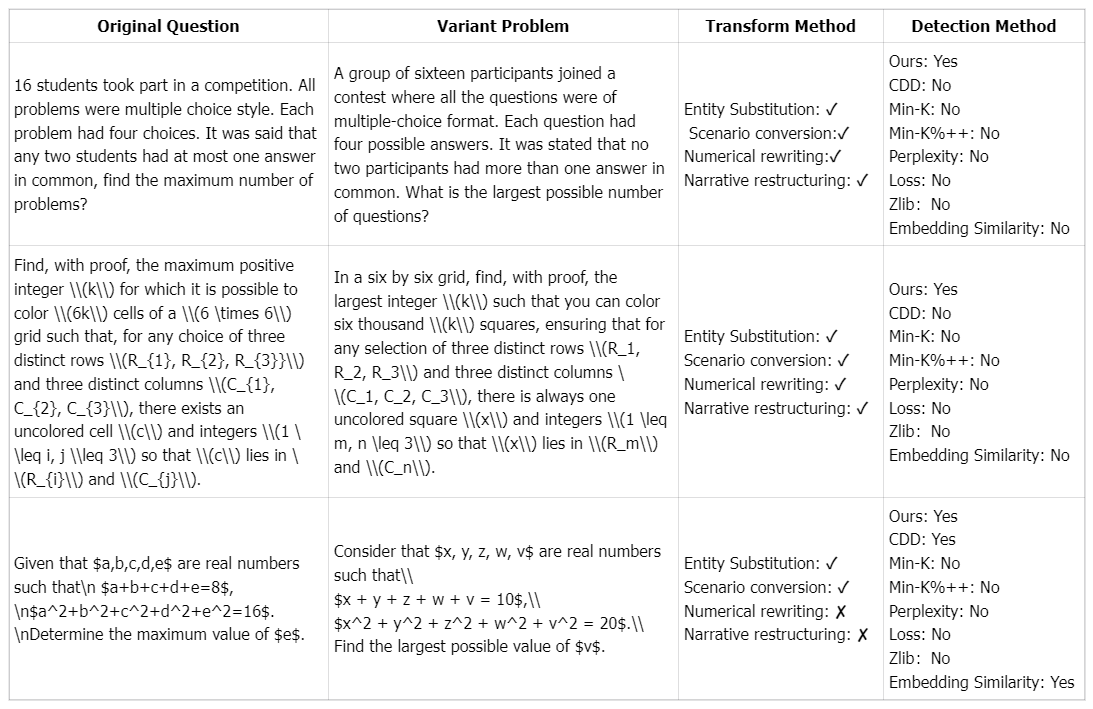}
    \caption{Compare the effectiveness of different detection methods on different variants}
    \label{fig:case_study}
\end{figure*}

The three representative cases examined above provide a mechanistic explanation for the macroscopic performance trends observed in Figure \ref{fig:case_study}. 
They demonstrate that the effectiveness of a detection method is not arbitrary but is determined by the intrinsic alignment between its underlying mechanism and the nature of the contamination. 
The superior performance of our DVD method stems from its unique capacity to probe the model's internal "cognitive state," enabling it to penetrate surface-level textual variations and identify the essential signal of memorization.

\subsubsection{Deeply Transformed Variants} Cases 1 and 2 represent deeply transformed variants where all four transformation methods (entity substitution, scenario conversion, etc.) are applied. Although the surface forms are completely different, such as changing "students" to "participants" and completely altering the narrative order, the core semantics remain consistent.

\textbf{Sampling and Output-Matching-Based Methods} , such as CDD, perform poorly is that they rely on surface form matching and fail to effectively capture deep logical equivalence. In Case 1, there is a significant surface difference between the original and variant problems, such as from "16 students took part in a competition" to "A group of sixteen participants joined a contest." Although the core semantics of the two problems remain consistent, the surface text changes significantly. The CDD method detects based only on surface symbol similarity, and when faced with surface-level changes (such as replacing some vocabulary or altering the narrative structure), CDD fails to recognize the deep similarities between these problems. CDD misjudges this surface difference as "output inconsistency," resulting in detection failure.

\textbf{Perplexity-Based Methods} perform poorly when faced with surface-level changes or structural perturbations, because they are highly sensitive to changes in the surface form of the text. These methods typically assess whether a model has memorized certain samples by measuring the perplexity of low-probability tokens. However, perplexity-based methods mainly rely on the model's confidence in generating known samples, neglecting the deeper semantics of the text. In Case 2, when the original problem is changed in variant names and narrative order (e.g., replacing \(i, j\) with \(m, n\)), this change does not fundamentally alter the mathematical structure of the problem, but the perplexity method may mistakenly classify it as non-memorized due to the model's lower confidence in generating these changes.

\textbf{The DVD method} is successful by probing the internal cognitive state of the model during generation, rather than analyzing the output text. Despite the surface differences, the model has memorized the core logical template for solving these problem types. 
When generating answers, it exhibits high confidence at the key reasoning steps and final answer. 
This results in low and stable "constitutive difficulty" values across samples, leading to a high variance score. 
Thus, DVD effectively detects contamination by identifying the model's familiarity with the underlying mathematical structure, bypassing surface-level noise.

\subsubsection{Simply Transformed Variants}
Case 3 is a simple variant that involves only entity substitution and numerical rewriting.
The mathematical problem (an application of the Cauchy-Schwarz inequality) remains identical, with only the variable names and the scenario  changed.

\textbf{Sampling and Output-Matching-Based Methods}, CDD, success was achieved in Case 3 because of the high text and semantic similarity between the original and variant problems, with only a few differences in specific phrases and variant names.

\textbf{Perplexity-Based Methods} perform poorly when facing surface-level changes or structural perturbations. Among multiple methods, only Embedding Similarity successfully identified the contamination, highlighting their fragility. Changes in specific tokens (variables, numbers) are sufficient to alter the probability distribution. For example, changing \(a, b, c, d, e\) to \(x, y, z, w, v\), and changing 8 and 16 to 10 and 20. These specific token changes are enough to significantly alter the model’s computation of the probability distribution of the entire sequence. The model has seen \(a + b + c + d + e = 8\), but has not seen \(x + y + z + w + v = 10\), so it perceives the latter sequence as having a slightly lower probability.

\textbf{The DVD method} performs excellently in such a simple entity substitution scenario, further demonstrating that by probing the model’s internal cognitive state during generation, our method effectively identifies and captures deep logical structures and semantic consistency.

\begin{figure*}[htbp]
    \centering
    \includegraphics[width=\textwidth]{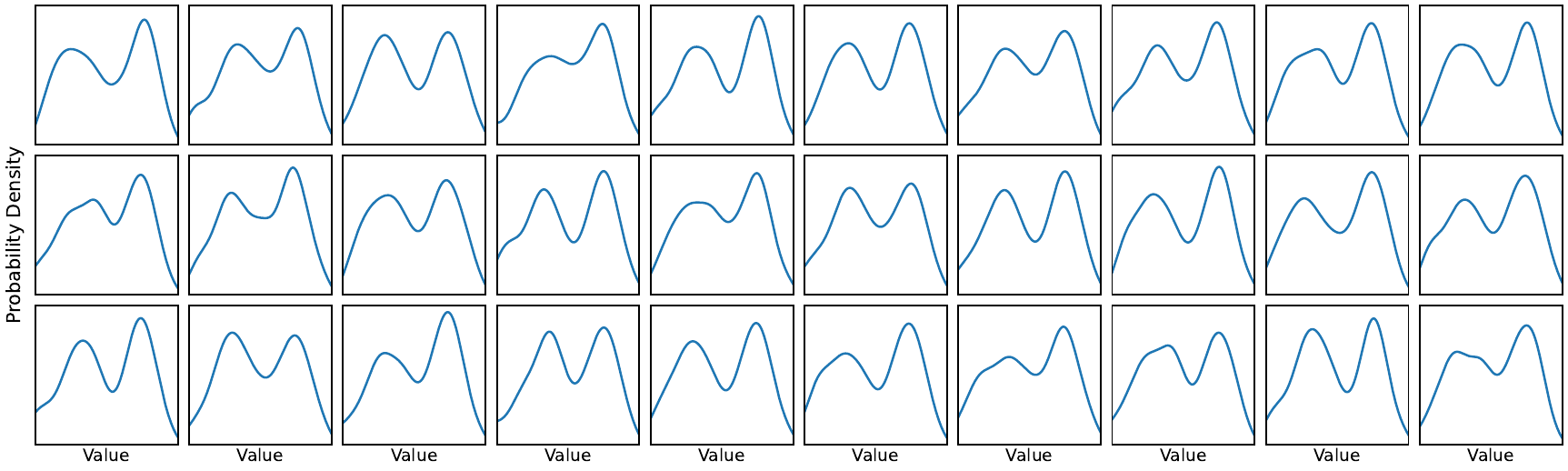}
    \caption{Probability distribution plot of the sum of log probabilities of the min-k tokens sampled from the distribution of contaminated samples}
    \label{fig:contaminated}
\end{figure*}

\begin{figure*}[htbp]
    \centering
    \includegraphics[width=\textwidth]{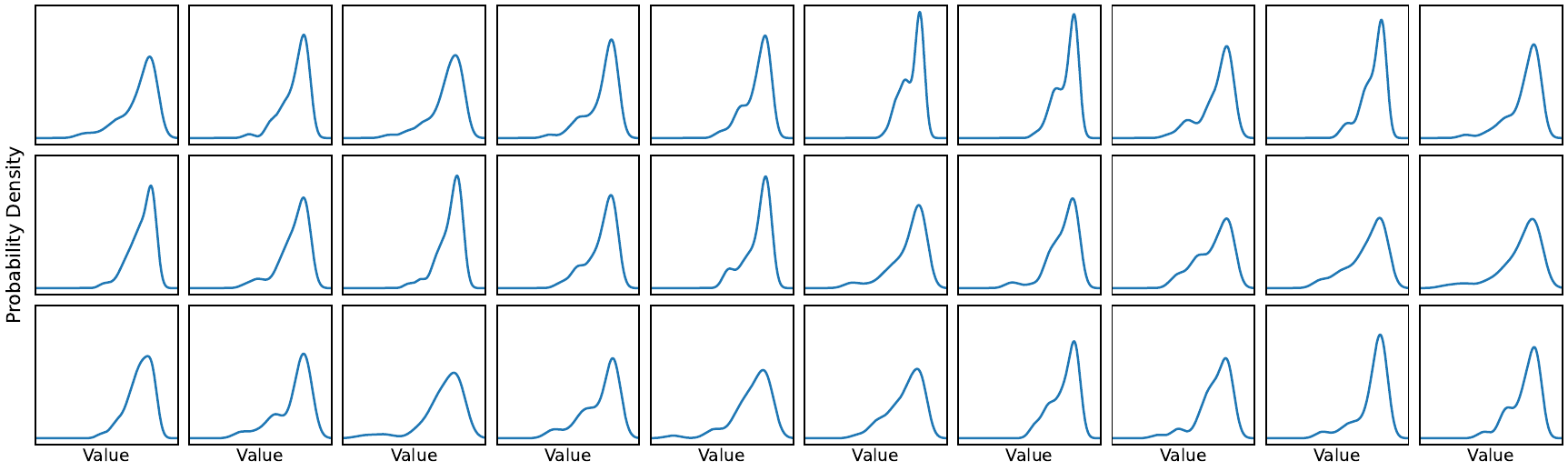}
    \caption{Probability distribution plot of the sum of log probabilities of the min-k tokens sampled from the distribution of uncontaminated samples}
    \label{fig:uncontaminated}
\end{figure*}

\begin{figure*}[!t]
    \centering
    \includegraphics[width=\textwidth]{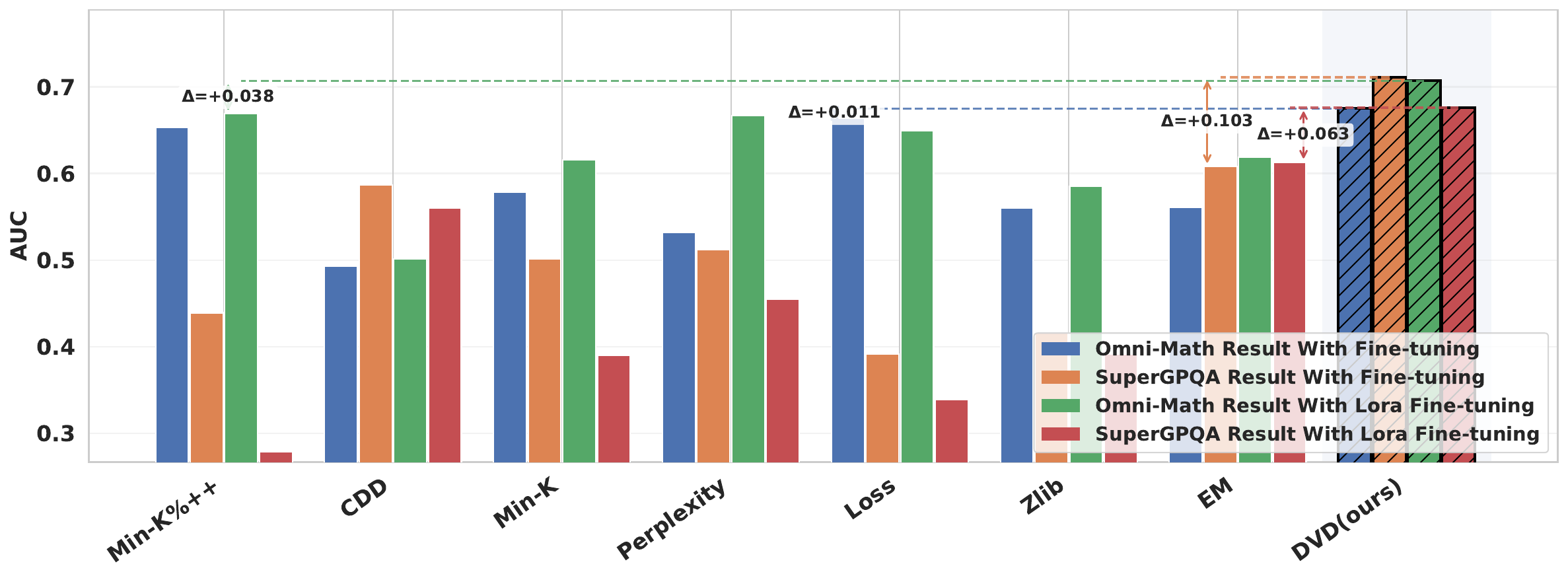}
    \caption{Performance of the DVD method on the Llama architecture compared to other baselines. EM denotes the Embedding-similarity method.}
    \label{fig:ablationStudyall}
\end{figure*}

\subsection{Statistical Evidence for Generation States: Memory Adherence and Perturbation Drift}
\label{sec:statistical_evidence}

This section provides the detailed statistical and experimental foundation for introducing the core generation states: \textbf{Memory Adherence} and \textbf{Perturbation Drift}. These states are not theoretical assumptions but stable, objectively observed modes of behavior resulting from a systematic statistical analysis of the model's generation process on contaminated and uncontaminated samples.

\subsubsection*{Experimental Setup}
To quantify the model's generation mechanism, we performed multiple repeated samplings for $100$ randomly selected samples (including both contaminated and clean examples) at a fixed sampling temperature $\tau$ (e.g., $\tau=0.8$). The primary statistical quantity analyzed is the distribution of the \textbf{log-likelihood of the sum of K-min token} at each generation step. This distribution characterizes the model's propensity to generate tokens with varying degrees of confidence and quality.

\subsubsection*{Bimodal Structure in Contaminated Samples}
As shown in \ref{fig:contaminated}, for samples subject to variant contamination, the distribution of the sum of K-min token log-likelihood consistently exhibits a \textbf{pronounced and repeatable bi-modal structure}. This characteristic structure is direct evidence that the model's generation process, when exposed to contamination, is not governed by a single random mechanism but dynamically switches between two distinct modes.

The analysis of the bi-modal structure reveals the following:
\begin{enumerate}
    \item \textbf{The First Peak (High-Confidence Region):} This peak is consistently located in the \textbf{higher} log-likelihood region. It corresponds to generation where the model selects high-probability, high-confidence tokens. This behavior indicates that the model is \textbf{adhering to answer fragments, linguistic patterns, or templates} encountered during training. We define this as the \textbf{Memory Adherence State}.
    \item \textbf{The Second Peak (Low-Confidence Region):} This peak resides in the \textbf{lower} log-likelihood region, corresponding to the selection of low-probability, high-randomness tokens. This mode suggests the model has \textbf{deviated from the memory track} and entered a more explorative, lower-confidence generation space, which we term the \textbf{Perturbation Drift State}.
\end{enumerate}
The bi-modal distribution directly proves that the model dynamically alternates between leveraging specific, strong memory structures and engaging in randomized, exploratory sampling on the same contaminated inputs.

\subsubsection*{Unimodal Structure in Uncontaminated Samples}
In stark contrast, uncontaminated (clean) samples, used as a control, consistently exhibit a \textbf{single, smooth, and approximately Gaussian distribution} (Figures \ref{fig:uncontaminated}).

The unimodal nature confirms that the model is following a \textbf{consistent, intrinsic random generation mechanism} without the disruptive influence of strong, pulling memory structures. The absence of a secondary peak supports the hypothesis that state-switching behavior is unique to contaminated data.

\subsection{Performance of the DVD method on the Llama architecture compared to other baselines.}
\label{sec:ablationStudyall}
Figures 
 \ref{fig:ablationStudyall} reports AUC on the Llama backbone across two benchmarks (Omni-Math and SuperGPQA) under both full fine-tuning and LoRA. Overall, DVD achieves the best or near-best performance in all settings and exhibits the most consistent gains. In contrast, baselines such as perplexity, Min-K\%, CDD, and Zlib vary substantially across datasets and fine-tuning regimes, indicating weaker robustness. These results suggest that DVD provides a more reliable detector of variant contamination and generalizes better across training setups.

\subsection{Significance testing experiments of the DVD method against other baselines}

\begin{table}[htbp]
\centering
\small
\setlength{\tabcolsep}{4pt}
\begin{tabular*}{\columnwidth}{@{\extracolsep{\fill}}lccc@{}}
\toprule
\textbf{Method} & \textbf{Omni-MATH} & \textbf{SuperGPQA} & \textbf{Conf} \\
\midrule
DVD (Ours) & $0.731 \pm 0.013$ & $0.700 \pm 0.016$ & -- \\
Min-K\%++ & $0.648 \pm 0.014$ & $0.364 \pm 0.018$ & 99\% \\
CDD & $0.513 \pm 0.012$ & $0.584 \pm 0.011$ & 99\% \\
Min-K & $0.531 \pm 0.015$ & $0.414 \pm 0.016$ & 99\% \\
Perplexity & $0.567 \pm 0.013$ & $0.404 \pm 0.017$ & 99\% \\
Loss & $0.567 \pm 0.012$ & $0.334 \pm 0.019$ & 99\% \\
Zlib & $0.550 \pm 0.014$ & $0.378 \pm 0.018$ & 99\% \\
EM & $0.544 \pm 0.011$ & $0.599 \pm 0.010$ & 99\% \\
\bottomrule
\end{tabular*}
\caption{Significance testing experiments of the DVD method against other baselines on Qwen2.5-7B-Instruct.}
\label{tab:qwen7b_significance}
\end{table}

Table~\ref{tab:qwen7b_significance} presents a comparison of the contamination detection performance of the proposed DVD method against various existing baselines on two challenging benchmarks, Omni-MATH and SuperGPQA, using the Qwen2.5-7B-Instruct model. The table also reports the confidence levels (column ``Conf'') from paired significance tests between DVD and each baseline. The results show that DVD significantly outperforms all competing methods on this model: it achieves an AUC of $0.731 \pm 0.013$ on Omni-MATH and $0.700 \pm 0.0016$ on SuperGPQA. The reported margins of error (standard deviations) are computed over 10 independent runs with different random seeds.

\subsection{Prompt}
\begin{figure}[htbp]
    \centering
    \includegraphics[width=\columnwidth]{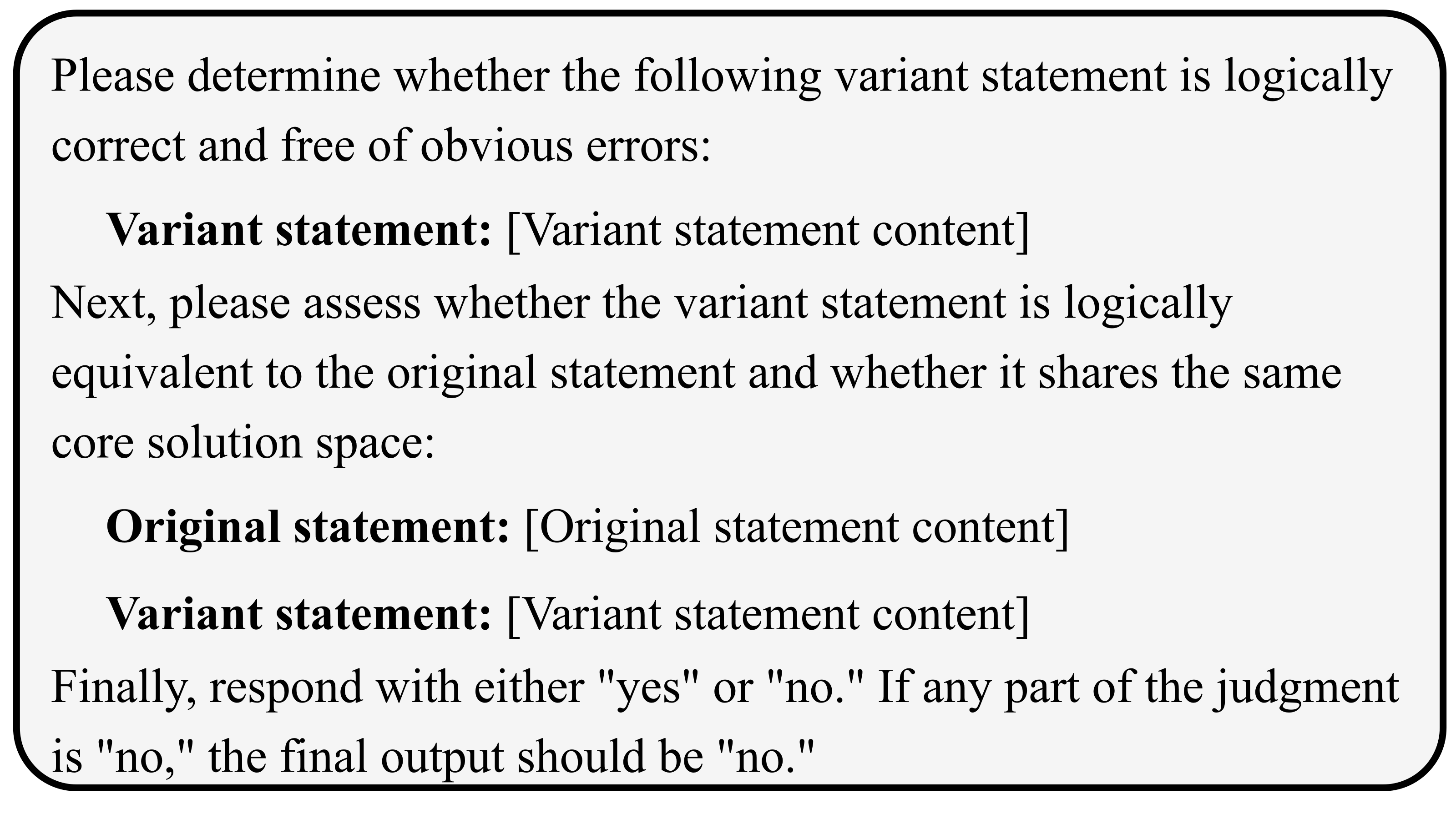}
    \caption{The prompt for rejection sampling}
    \label{fig:Reject_prompt}
\end{figure}

\begin{figure*}[htbp]
    \centering
    \includegraphics[width=\textwidth]{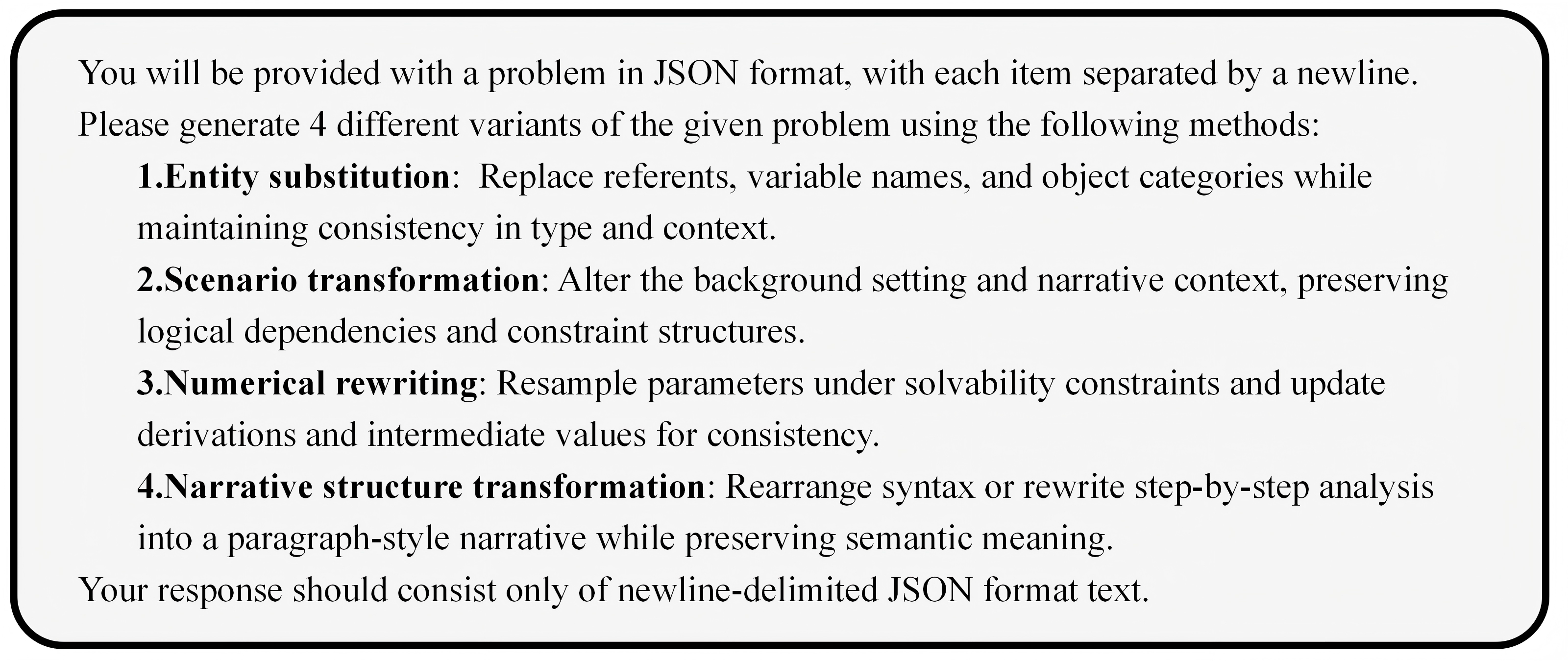}
    \caption{The prompt used to generate variations}
    \label{fig:Variant_prompt}
\end{figure*}

Figures~\ref{fig:Variant_prompt} and~\ref{fig:Reject_prompt} show the prompts used in our data construction pipeline: Figure~\ref{fig:Variant_prompt} is the prompt for generating semantic variations, and Figure~\ref{fig:Reject_prompt} is the prompt for rejection sampling.

\end{document}